\documentclass[10pt,twocolumn,letterpaper]{article}

\usepackage{wacv}
\usepackage{times}
\usepackage{epsfig}
\usepackage{graphicx}
\usepackage{amsmath}
\usepackage{amssymb}
\usepackage{tabularx}

\def\wacvPaperID{492} 

\wacvfinalcopy 

\ifwacvfinal
\def\assignedStartPage{9876} 
\fi
\ifwacvfinal
\usepackage[breaklinks=true,bookmarks=false]{hyperref}
\else
\usepackage[pagebackref=true,breaklinks=true,colorlinks,bookmarks=false]{hyperref}
\fi

\ifwacvfinal
\else
\fi

\begin{document}

\title{
We don't Need Thousand Proposals:\\ Single Shot Actor-Action Detection in Videos
}


\author{Aayush J Rana \\
{\tt\small aayushjr@knights.ucf.edu}
\and
Yogesh S Rawat\\
{\tt\small yogesh@crcv.ucf.edu}
\and
\and
Center for Research in Computer Vision\\
University of Central Florida\\
Orlando, FL, 32816\\
}

\maketitle


\begin{abstract}

We propose \textit{SSA2D}, a simple yet effective end-to-end deep network for actor-action detection in videos. The existing methods take a top-down approach based on region-proposals (RPN), where the action is estimated based on the detected proposals followed by post-processing such as non-maximal suppression. While effective in terms of performance, these methods pose limitations in scalability for dense video scenes with a high memory requirement for thousands of proposals. We propose to solve this problem from a different perspective where we don't need any proposals. \textit{SSA2D} is a unified network, which performs pixel level joint actor-action detection in a \textit{single-shot}, where every pixel of the detected actor is assigned an action label. \textit{SSA2D} has two main advantages: 1)  It is a fully convolutional network which does not require any proposals and post-processing making it \textit{memory} as well as \textit{time efficient}, 2) It is easily \textit{scalable} to dense video scenes as its memory requirement is independent of the number of actors present in the scene. We evaluate the proposed method on the Actor-Action dataset (A2D) and Video Object Relation (VidOR) dataset,  demonstrating its effectiveness in multiple actors and action detection in a video. \textit{SSA2D} is \textit{11x faster} during inference with comparable (\textit{sometimes better}) performance and \textit{fewer} network parameters when compared with the prior works. Code available at \url{https://github.com/aayushjr/ssa2d}
  
\end{abstract}

\section{Introduction}
\label{sec:intro}

Actor-action detection in videos is a challenging problem where the goal is to detect all the actors in the video and determine which different actions they are performing. One natural solution to this problem is to perform object detection and identify all the actors and classify those detected actors for corresponding actions. Motivated by this, existing methods utilize region proposal (RPN) \cite{frcnn} based approach \cite{dang2018actor, A2D2018stanford}, where they first detect proposal for objects and use them for action detection.

\begin{figure}
\begin{center}
\includegraphics[width=0.48\textwidth]{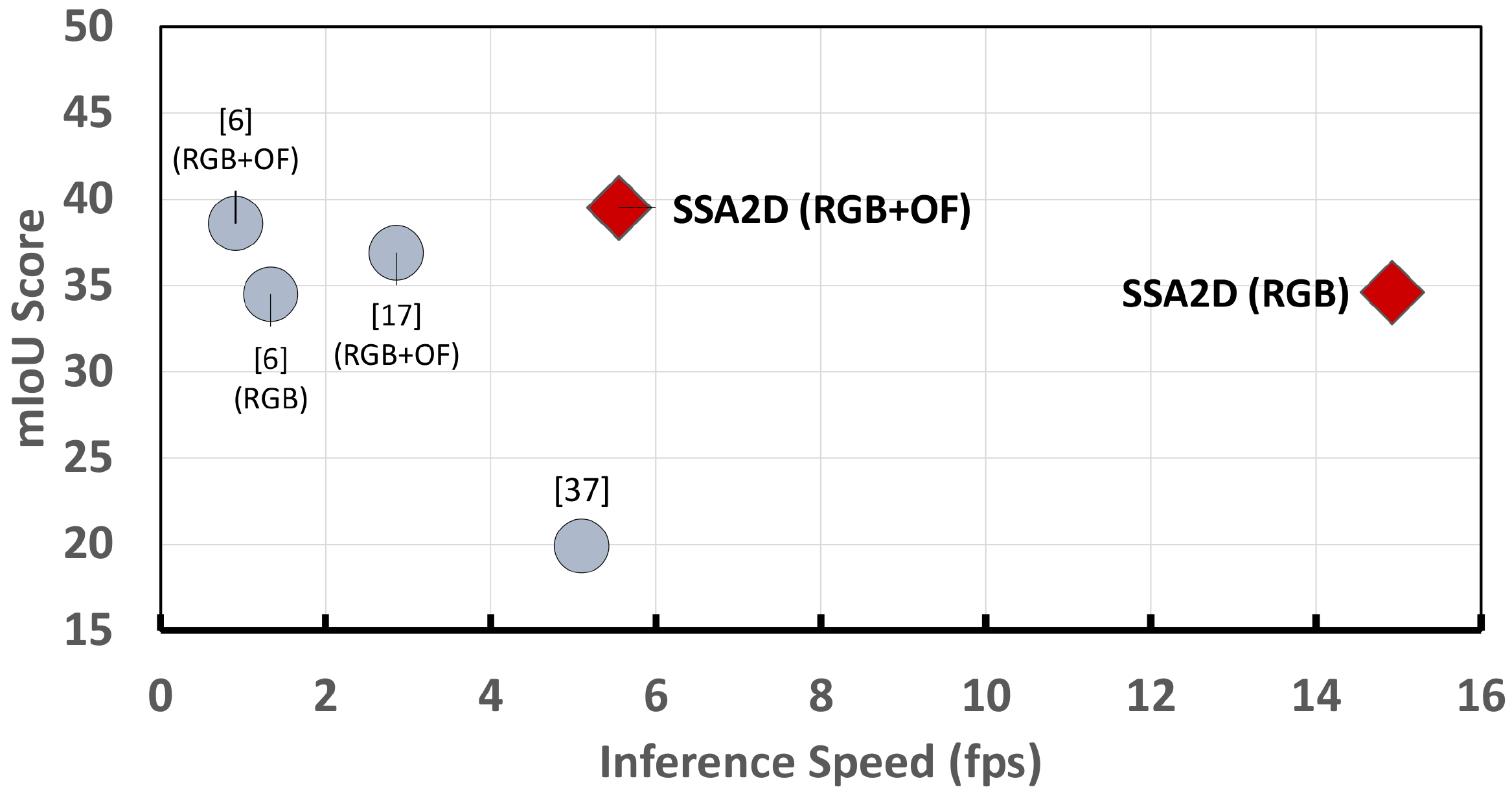}
\end{center}
\caption{ A comparative analysis of SSA2D with existing methods in terms of speed and performance. We observe that SSA2D is faster with comparable performance. The x-axis represents inference speed in frames per second and the y-axis represents the mean pixel-wise intersection over union score for joint actor-action detection. (RGB+OF - Using both RGB and optical flow).
} 
\label{fig:intro}
\end{figure}

Region proposal and ROI pooling based methods \cite{frcnn, girshick2015fast} have been shown to be extremely successful for object detection in images. The approach usually involves detection of thousands of proposals followed by a post-processing cleanup using non-maximal suppression. In videos, we have an additional time dimension, and detecting proposals for every frame can be \textit{memory intensive}. Therefore, the existing methods \cite{dang2018actor, A2D2018stanford} relying upon RPN approach can perform actor-action detection only on \textit{single frame} at a time. Also, with multiple actors and multiple actions per actor in a scene, these approaches become complex and inefficient as thousands of region proposals per frame are required for training. Due to these limitations, such networks have to be trained in multiple stages \cite{dang2018actor}, leading to an increase in the training and inference time. Single shot detection methods~\cite{ssd,tian2019fcos,yolo,yolo9000} overcome such region proposal limitations, but they don't translate well for video actor-action detection.

In this work, we present an alternative approach which is simple yet effective and overcomes these limitations. We propose SSA2D, a novel end-to-end deep network which does not require proposals and performs \textit{single shot} actor-action detection in videos. SSA2D is an encoder-decoder based unified network, which utilizes spatio-temporal contextual information between objects and their surrounding pixels for joint detection of \textit{multiple} objects and activities in \textit{multiple} input video frames at once. In RPN based approach, \textit{ROI-Pooling} \cite{frcnn} is used to extract object focused features for each proposal and therefore it leads to a higher memory consumption with increase in number of objects in the scene. In contrast, we propose \textit{Single-Shot Attentive Masking (SSA-Masking)}, which utilize a spatio-temporal unified mask (\textit{STU-Mask}) for all the objects to perform feature filtering at once. This unified masking makes the proposed approach much more efficient for dense video scenes. 

A prior knowledge of actor presence provides crucial information regarding the performed action. A joint training of actor and action detection allows us to utilize actor-prior (\textit{A-Prior}) to assist in action detection. We propose actor-prior infusion (\textit{AP-Infusion}), which make use of spatio-temporal actor features as actor-priors. We demonstrate that \textit{AP-Infusion} helps in improving performance of action detection as well as actor detection. Apart from this, SSA2D utilize atrous convolutions \cite{chen2017atrous} and feature pyramid network \cite{lin2017fpn}, which we adapt for videos, to address the issue of \textit{multiple scales} of actors present in a scene. SSA2D is jointly trained end-to-end for three objectives, actor detection, action detection, and STU-Mask, with the help of a combination of pixel-wise cross-entropy loss and a generalized dice-loss \cite{sudre2017generaliseddiceloss}, which we extend for videos as Generalized 3D Dice Loss to balance the foreground and the background pixels in videos.
 

The proposed SSA2D model has several advantages over existing methods: 1) It requires fewer network parameters making it memory efficient, 2) It has faster inference and also requires less training time, 3) The memory requirement of SSA2D does not depend upon the number of actors present in the video, and therefore it does not have any scalability issues for dense scenes, and finally 4) SSA2D can perform spatio-temporal actor-action detection on multiple input video frames at once making it more efficient when compared with existing approaches which can perform detection only on single frame at a time. 
We make the following contributions in this work:
\begin{itemize}
\item We present SSA2D, a novel end-to-end network, to perform pixel-wise detection of actors and actions in videos. SSA2D is trained jointly to detect multiple actors and their actions simultaneously, while not relying on region proposals or any external post-processing.
\item We propose SSA-Masking, which utilize unified spatio-temporal mask for selective feature extraction of multiple actors at once. In contrast to ROI-Pooling, it is much more efficient for dense video scenes.
\item We propose the use of actor-prior for spatio-temporal pixel-wise action detection with the help of spatio-temporal actor features, leading to an improved performance of both action as well as actor detection in joint learning.
\end{itemize}
We perform extensive experiments on A2D and VidOR dataset and demonstrate SSA2D to be efficient as well as effective for joint actor-action detection in videos. The proposed method is significantly efficient in inference ($\sim$11x faster for RGB and $\sim$6x faster for RGB+optical-flow) as well as in training (2x faster than state-of-the-art) with fewer network parameters and yet achieve performance comparable (sometimes even better) to the prior works.

\begin{figure*}[ht!]
\begin{center}
\includegraphics[width=0.95\linewidth]{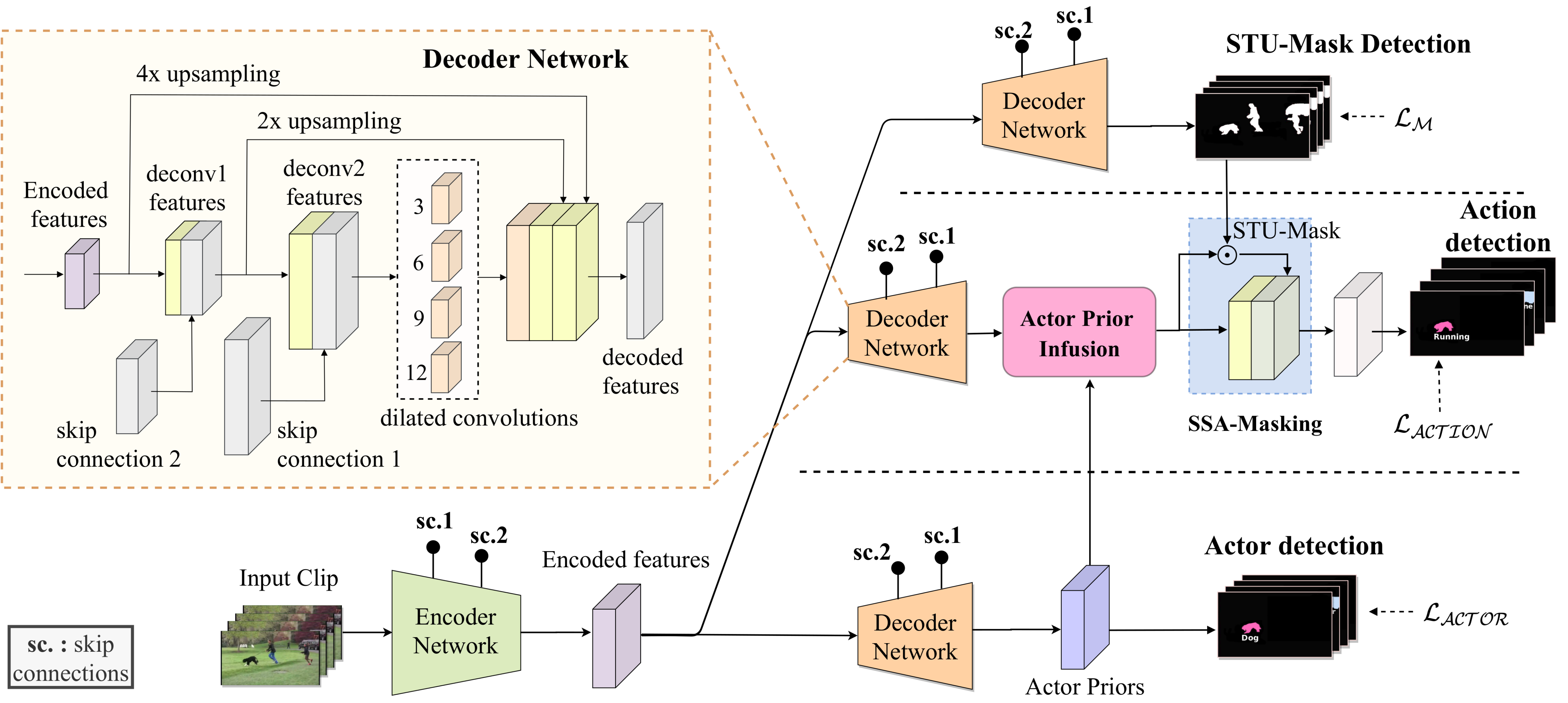}
\end{center}
\vspace{-0.5em}
\caption{Overview of the proposed method for pixel-wise actor-action detection showing the overall SSA2D architecture. The 3D convolution based encoder network extracts features which are used in three separate branches for actor, action, and  \textit{STU-Mask} detection. The action detection branch use \textit{AP-Infusion} to utilize actor-priors and \textit{SSA-Masking} to focus on activity regions in the video. All three branches use a decoder network which has similar architecture, however the weights are not shared among these tasks.
}
\label{fig:adrec_cls}
\vspace{-1.5em}
\end{figure*}

\section{Related work}


\subsection{Human action detection in videos}
Action detection in videos require spatio-temporal localizations of actors in each frame which is then used for classification. Extending the image classification models ~\cite{frcnn, yolo, yolo9000, ssd, gkioxari2017detecting}, prior CNN based works detect humans in each frame and combine them temporally to form action tubes while classifying at clip level~\cite{yang2017spatio, peng2016multi, yang2019step, li2018recurrent}, leveraging existing classification techniques from~\cite{zisserman2017quo, hara2018resnet3d, hou2017tcnn, qiu2017p3d, xie2018rethinking, tran2015c3dfb, simonyan2014twoStream, vyas2020multi}.~\cite{hou2017tcnn} uses RPN based approach to detect humans in each frame and then forms action tubes by stitching them together, followed by Tube of Interest (TOI) pooling and action classification.~\cite{hou2017end} does TOI pooling based on foreground segmentation map from an encoder-decoder based network.~\cite{girdhar2018actionTransformer} uses RPN along with transformer based attention map that detects and classifies actions.~\cite{duarte2018videocapsulenet} uses a 3D capsule based CNN, where the authors apply routing-by-agreement algorithm to capture various action representations, leading to localize actions and classify them at the same time. Although prior works show great improvements on action detection in videos, they are limited by complex region proposal network coupled with region pooling or can only detect and classify single actor per video, creating a challenge to adapt it to denser real-life scenarios.

\subsection{Actor and action detection in videos}
Actor-action detection problem is related to identifying the actors and their corresponding actions in a given clip, where both semantic localization and classification is required. The authors in~\cite{A2DCVPR2015} proposed the A2D dataset, a large scale benchmark dataset to study actor-action detection problem. An early approach of adaptive grouping of segments during inference improves segmentation in A2D~\cite{A2D2016GPM}.~\cite{yan2017weakly} proposed weakly supervised method and train the model using only video-level tags. A two-stage model was proposed by ~\cite{A2D2017TSMT}, where  objects are detected first and their bounding box are refined for segmentation outputs.~\cite{gavrilyuk2018actor,mcintosh2020visual} use sentence priors to detect actor-actions in videos. Authors in~\cite{A2D2018stanford} propose a joint end-to-end model which uses two-stream input (RGB + optical flow) to classify object regions and perform segmentation on them. Conceptually based on~\cite{he2017mask}, they generate semantic features and use RPN to segment and classify actor-action pairs.~\cite{dang2018actor} use similar approach and propose segmentation based region proposal and pooling to detect actor and action classes. They apply a region pooling based fully convolutional segmentation network for their actor segmentation, followed by 2D ResNet-101~\cite{he2016deep} for action classification. Although prior works show great improvements on joint actor-action classification, they are limited by expensive region proposal and pooling which increases the approach's complexity. 


\section{Proposed method}


Given a video $V \in \mathbb{R}^{T \times H \times W \times 3}$ as input, SSA2D jointly predicts actor detection $ActorD\in \mathbb{R}^{T \times H \times W \times C_{actor}}$ and action detection $ActionD\in \mathbb{R}^{T \times H \times W \times C_{action}}$. Here, $T$ is the number of frames in the input clip, $H$ is the height and $W$ is the width of the video frames, $C_{actor}$ is total number of actor classes, and $C_{action}$ is the total number of action categories. In addition to these two, SSA2D also predicts a spatio-temporal mask $STUMask \in \mathbb{R}^{T \times H \times W \times 2}$, which is used for SSA-Masking. SSA2D consists of an encoder network $E$, and three separate branches for actor, action, and STU-Mask detection. Each task utilize a decoder network ($D$), which has similar architecture for all the three branches. An overview of SSA2D is shown in Figure \ref{fig:adrec_cls}.

\subsection{Encoder network}

Understanding and extracting relevant features both spatially and temporally is crucial in learning a video’s actor-action relations. We utilize a \textit{3D convolution} based encoder $E$ that extracts actor and action related feature volume $f_{enc}$ from a given input video clip $V \in \mathbb{R}^{T \times H \times W \times 3}$. The network takes a video clip as input with [$T\times H\times W \times 3$] with $T$ frames at a resolution of $H\times$W and outputs a feature volume $f_{enc} \in \mathbb{R}^{\frac{T}{4}\times \frac{H}{16}\times \frac{W}{16}}$ as video encodings. We use I3D \cite{zisserman2017quo} model as our encoder where we adapt the network by controlling the pooling strides (more details in supplementary). This encoder can use any state-of-the art 3D convolution network. 


\subsection{Decoder network}
\label{sec:decoder}
The spatio-temporal features $f_{enc}$ extracted using the encoder network $E$ needs to be decoded into a larger fine-grained resolution for jointly detecting the actors and actions. 
The decoder network $D$ takes $f_{enc}$ as input and performs a series of 3D deconvolution~\cite{deconvolutionnetworks} and upsampling operations to get the desired resolution for fine-grained pixel-wise detection. We upsample the encoded features to [$\frac{T}{2}\times \frac{H}{4}\times \frac{W}{4}$] as a final resolution to reduce parameters. We add skip connections from the encoder network to every deconvolution layer to preserve the suppressed features during downsampling. Adapting multi-scale object feature learning techniques from images, we extend atrous convolutions ~\cite{chen2017atrous, chen2018encAtrous} and feature pyramid network~\cite{lin2017fpn, kirillov2019panopticfpn} to 3D architecture for videos. Atrous convolutions helps encode multi-scale contextual information around each pixel while feature pyramid helps preserve features of smaller objects.

The same decoder architecture $D$ is used in all the three branches, however, the network parameters are not shared as these branches solve different tasks. The final output from all the branches is upsampled to match the resolution of the input video with the help of linear interpolation. A detailed architecture of $D$ is shown in Figure~\ref{fig:adrec_cls} and more details are provided in the supplementary.

\subsection{Actor detection}
For pixel-wise actor detection, the actor detection branch utilizes encoded video features $f_{enc}$ and learns pixel-wise $C_{ap}$ actor prior (\textit{A-prior} $\in \mathbb{R}^{T \times H \times W \times C_{ap}}$) with the help of a decoder network $D_{actor}$. A final 3D convolution layer takes the learned \textit{A-Prior} and predicts $C_{Actor}$ channels for each pixel ($C_{Actor}$ being the total number of actors present in the dataset). A \textit{Softmax} activation is applied across actor channels for each pixel location as each pixel will correspond to only one of the actors. This gives us $ActorD$ for pixel-wise actor detection in the input video. The scores in each channel corresponds to one of the actor class and indicates its presence in that spatio-temporal location.

\subsection{Action detection}
The action detection branch $Action$ takes the spatio-temporal features $f_{enc}$ from the encoder network $E$ as input and uses the decoder architecture $D_{action}$ from section \ref{sec:decoder} to learn action relevant feature maps $f_{a}$ for action detection. As each actor's interaction with surrounding objects is decisive in inferring its actions, the actor detection branch will have more meaningful features corresponding to each actor. However, it is also important to focus only on the spatio-temporal region where the action is occurring. To address these issues, we propose Actor Prior Infusion (\textit{AP-Infusion}) and Single-Shot Attentive Masking (SSA-Masking), which allow the network to filter and learn meaningful interaction between the detected actors for action detection.


\paragraph{\textbf{Actor Prior Infusion (\textit{AP-Infusion})}}
The Actor Prior Infusion (\textit{AP-Infusion}) provides additional information to the action detection network in form of latent actor representations. This is done by integrating \textit{A-priors} with action related features, adding more actor focused contextual information and helps in action detection. As shown in Figure~\ref{fig:adrec_cls}, the \textit{A-priors} $f_{ap}$ are integrated with action features $f_a$ from the decoder network in $Action$ branch as $f_{act} = Conv3D(<f_a, f_{ap}>)$, 
where $Conv3D$ is 3D convolution operation and $<>$ represented feature concatenation. We also experimented with feature addition and observed similar performance.



\paragraph{Single-Shot Attentive Masking (\textit{SSA-Masking}):}
Instead of generating proposal boxes from external networks~\cite{frcnn} or using all possible region boxes~\cite{yolo}, we use single-shot attentive masking for feature filtering. A fine-grained spatio-temporal region is helpful to filter and improve the coarse actor-action detection results. To get this spatio-temporal mask, the features $f_{enc}$ from the encoder network $E$ are passed to a decoder network $D_{STU-Mask}$ which predicts pixel-wise scores $STU-Mask\in \mathbb{R}^{T \times H \times W \times 2}$ for each spatio-temporal location in the input video. Each pixel's score in the \textit{STU-Mask} indicates whether it is relevant to the action or not. The network learns to identify potential actor regions through the \textit{STU-Mask}. This mask from the $D_{STU-Mask}$ is used as spatio-temporal unified mask $f_{mask} \in \mathbb{R}^{T \times H \times W \times 1}$ to filter the spatio-temporal features for action detection. 

The action features $f_{act}$ augmented with actor-priors are filtered using \textit{SSA-Masking}. The augmented features $f_{act}$ are integrated with the \textit{STU-Mask} [$f'_{act}$ = \textit{$f_{act}$} $\odot$ \textit{$f_{mask}$}] to get the filtered features $f'_{act}$. The filtered features $f'_{act}$ are integrated back with the original action features $f_{act}$ [$f''_{act} = <f'_{act}, f_{act}>$] to keep both action as well as contextual background features for an effective learning. With this masking, forward pass only learns detection of useful feature regions while backward pass has minimal gradient update for unrelated regions. Furthermore, the \textit{SSA-Masking} can be done on the whole frame in a single-shot, removing the need for extracting multiple region proposal boxes and performing ROI/TOI pooling. The masking can be done within the network, making this an end-to-end architecture. During training, we use the ground truth \textit{STU-Mask}. While testing, we extract the $D_{STU-Mask}$ detection results and pass that as the \textit{STU-Mask} within the network. Finally, the output feature from \textit{SSA-Masking} is used to predict $ActionD$ with $C_{Action}$ channels using 3D convolution, where each channel corresponds to one action class. Since each pixel is evaluated individually, it can be formulated to have multi-labels and multi-class predictions.


\subsection{Objective function}
The proposed network is trained end-to-end with joint learning of three tasks: actor detection, action detection, and STU-Mask detection. Since we predict pixel-wise maps for each branch, we have to consider the large imbalance in active and non-active pixels, with fewer active pixels for sparse scenes. This imbalance is handled using ratio loss for the scene. In case of image segmentation, this can be computed as a ratio of foreground pixels to background using the Generalized Dice Loss \cite{sudre2017generaliseddiceloss}. We extended it to videos as Generalized 3D Dice Loss with the following formulation:
\begin{equation}
\small
\label{eq:dice_loss}
\begin{split}
    \mathcal{L_{DL}} = 1 - \frac{2\sum_{c=1}^{C}\sum_{i=1}^{N}p_{ci}*\hat{p_{ci}}}{\sum_{c=1}^{C}(\sum_{i=1}^{N}p_{ci}^2 + \sum_{i=1}^{N}\hat{p_{ci}}^2 + \epsilon)}
\end{split}
\end{equation}

where the dice coefficient score is computed per class $C$ of given task, $N$ is total number of pixels in segmentation map of a video clip, probability $p_{ci} \in (0,1)$ is the ground-truth segmentation map, and $\hat{p_{ci}} \in (0,1)$ is the network's predicted segmentation map probability.

The actor detection loss is defined as the negative log-likelihood of the ground truth class and is computed as categorical cross-entropy per pixel. For $C_{actor}$ set of actor classes, the actor detection head generates $C_{actor}$ segmentation maps, where each pixel's ground truth actor class is $x$ and predicted actor class is $\hat{x}$. The loss is calculated for each pixel across all classes and then averaged over all pixels, which gives us the following loss formulation:
\begin{equation}
\small
\label{eq:actor_loss}
\begin{split}
    \mathcal{L_{ACTOR}} = [-\frac{1}{N} \sum_{i=1}^{N} \sum_{j=1}^{C_{actor}} (x_{i,j})log(\hat{x}_{i,j})] + \mathcal{L_{DL}}.
\end{split}
\end{equation}

Action detection is also defined similarly to actor detection, with $C_{action}$ segmentation maps generated. For each pixel's ground truth action class $y$ an action class $\hat{y}$ is predicted, and the loss is:
\begin{equation}
\small
\label{eq:action_loss}
\begin{split}
    \mathcal{L_{ACTION}} = [-\frac{1}{N} \sum_{i=1}^{N} \sum_{j=1}^{C_{action}} (y_{i,j})log(\hat{y}_{i,j})] + \mathcal{L_{DL}}.
\end{split}
\end{equation}

We look at the STU-Mask detection task as a binary segmentation task, where all the actor pixels are considered as positive and all others as negative. The loss is computed using binary cross-entropy in combination with the dice loss:
\begin{equation}
\small
\label{eq:fg_loss}
\begin{split}
    \mathcal{L_{M}} = [-\frac{1}{N} \sum_{i=1}^{N}p_ilog(\hat{p_i}) - (1-p_i)log(1-\hat{p_i})] + \mathcal{L_{DL}},
\end{split}
\end{equation}

where $\hat{p_i}$ is the prediction and $p_i$ is the ground-truth. The total loss is a combination of these losses and is defined as:
\begin{equation}
\small
\label{eq:final_loss}
\begin{split}
    \mathcal{L} = \mathcal{L_{ACTOR}} + \mathcal{L_{ACTION}} + \mathcal{L_{M}}.
\end{split}
\end{equation}



\begin{table*}[t!] 
\begin{center}
\footnotesize
\begin{tabularx}{\textwidth}{c|c||@{\extracolsep{\fill}}c c c||c c c||c c c||c}
\hline
Input & Method & \multicolumn{3}{c||}{Actor} & \multicolumn{3}{c||}{Action} & \multicolumn{3}{c||}{Joint (A,A)} & Time (ms)\\
& & glo           & ave           & mIoU          & glo           & ave           & mIoU          & glo           & ave           & mIoU          & per frame   \\
\hline
& GPM + TSP ~\cite{A2D2016GPM} & 85.2          & 58.3          & 33.4          & 85.3          & 60.5          & 32.0          & 84.2          & 43.3          & 19.9          & -           \\
& GPM + GBH ~\cite{A2D2016GPM} & 84.9          & 61.2          & 33.3          & 84.8          & 59.4          & 31.9          & 83.8          & 43.9          & 19.9          & -           \\
RGB & Chen et al. ~\cite{chen2020learning}* & 91.3         & 49.16         & 49.2          & 87.44     & 35.12     & 38.7      & 87.1        & 43.06     & 26.7      & -     \\
& Ji et al. ~\cite{A2D2018stanford}  & 93.7          & 79.5          & 66.5          & 86.3          & 60.4          & 36.8          & 87.8          & 46.2          & 29.4          & -           \\
& Dang et al. ~\cite{dang2018actor} & 95.0 & 85.5          & 67.0          & 92.9 & 68.8          & 48.1 & 92.5 & 51.5          & 34.5          & 750         \\
& SSA2D (Ours) & 96.1          & 79.4 & 66.8 & 94.4          & 66.2 & 46.5          & 93.8          & 49.3 & 34.6 & 67 \\ 
\hline \hline

& TSMT + GBH ~\cite{A2D2017TSMT}& 85.8           & 72.9          & 42.7          & 84.6          & 61.4          & 35.5          & 83.9          & 48.0          & 24.9          & -            \\
RGB & TSMT + SM ~\cite{A2D2017TSMT}& 90.6           & 73.7          & 49.5          & 89.3          & 60.5          & 42.2          & 88.7          & 47.5          & 29.7          & -            \\
+ & Gavrilyuk et al. ~\cite{gavrilyuk2018actor}\textsuperscript{\rm \dag} & 92.8      & 71.4      & 53.7      & 92.5      & 69.3      & 49.4      & 91.7     & 52.4      & 34.8      & - \\
OF & Ji et al. ~\cite{A2D2018stanford}& 94.5           & 79.1          & 66.4          & 92.6          & 62.9          & 46.3          & 92.5          & 51.4          & 36.9          & 350**            \\
& Dang et al. ~\cite{dang2018actor}& 95.3           & 86.0 & 68.1 & 93.4          & 70.7 & 51.1 & 93.0            & 56.4 & 38.6 & 1100         \\
& SSA2D (Ours) & 96.2 & 80.1          & 67.5          & 94.9 & 69.1          & 51.3        & 95.0 & 54.7          & 39.5 & 180 \\
\hline
\end{tabularx}
\end{center}
\caption{Quantitative comparison of SSA2D on A2D dataset with prior approaches using RGB and RGB + optical flow as input, reporting average per-class accuracy $(ave)$, global pixel accuracy $(glo)$ and mean pixel Intersection-over-Union $(mIoU)$ for each task. \textsuperscript{\rm \dag} Uses sentence priors. \textit{*Uses weakly-supervised training.} ** Is time adjusted for same hardware setting by correspondence with authors. 
}
\label{table_A2D_RGB}
\end{table*}

\begin{table*}[t!]
\begin{center}
\small
\begin{tabularx}{\textwidth}{c|@{\extracolsep{\fill}}c||c c c|c c c|c c c}
\hline
Dataset & Method & \multicolumn{3}{c|}{Actor} & \multicolumn{3}{c|}{Action} & \multicolumn{3}{c}{Joint (A,A)}\\
& & glo & ave & mIoU & glo & ave & mIoU & glo & ave & mIoU\\
\hline
A2D & \textbf{Full (RGB only)} & \textbf{96.2} & \textbf{80.1} & \textbf{67.5} & 94.4 & \textbf{66.2} & \textbf{46.5} & \textbf{93.8} & \textbf{49.3} &  \textbf{34.6} \\
A2D & w/o \textit{Actor-Prior} & 96.1 & 79.1 &  65.7 & 93.9 & 61.5 & 40.9 & 93.4 & 46.3 &  32.1\\
A2D & w/o \textit{SSA-Masking} & 96.1 & 79.9 & 67.2 & \textbf{94.6} & 63.9 & 43.6 & 93.7 & 48.2 &  33.8\\
A2D & w/o atrous convolutions & 92.4 & 76.4 & 62 & 94.1 & 62.3 & 41.6 & 92.8 & 45.2 &  31.7\\
A2D & w/o multi-scale & 96.0 & 79.8 & 66.5 & 94.2 & 63.8 & 43.1 & 93.6 & 47.9 &  33.1\\
\hline
\hline
VidOR & \textbf{Full (RGB only)}    &   \textbf{72.2}          & \textbf{7.6}          & \textbf{5.1}          & \textbf{66.8}          & \textbf{33.2}          & \textbf{7.9}          & \textbf{41.7}          & \textbf{15.7}          & \textbf{2.1}  \\
& & (54.1) & (20.5) & (12.5) & (70.2) & (40.8) & (11.7) & (44.2) & (18.8) & (5.1) \\
VidOR & w/o \textit{Actor-Prior} & 71.1 & 7.1 & 4.1 & 61.8 & 28.3 & 5.9 & 37.7 & 12.1 & 1.1 \\
VidOR & w/o \textit{SSA-Masking} & 71.8 & 7.1 & 4.3 & 65.4 & 31.7 & 7.1 & 39.2 & 15.2 & 1.8 \\
VidOR & w/o atrous convolutions & 71.5 & 6.4 & 3.4 & 63.1 & 30.8 & 6.2 & 38.3 & 14.4 & 1.3 \\
VidOR & w/o multi-scale & 71.2 & 6.1 & 3.1 & 61.7 & 29.5 & 6.0 & 37.8 & 14.1 & 1.1 \\ 
\hline
\end{tabularx}
\end{center}
\caption{Ablation study of various components of SSA2D and their effect on actor-action detection on A2D and VidOR dataset. We report scores on average per-class accuracy $(ave)$, global pixel accuracy $(glo)$ and mean pixel Intersection-over-Union $(mIoU)$. Values in bracket represent scores for the 20 most frequent classes.
}
\label{table_A2D_ablation}
\end{table*}


\subsection{Implementation and training details}
We implement the proposed method in Keras~\cite{chollet2015keras} with Tensorflow backend. The encoder block uses I3D~\cite{zisserman2017quo} pre-trained on Kinetics-400. We input a video clip of temporal resolution (T) of 16 frames and spatial resolution 224 x 224. The final output of the encoder network is 4 x 14 x 14, which we then upsample to 8 x 112 x 112 for \textit{STU-Mask} detection branch and 8 x 56 x 56 for actor and action detection branch. For the RGB + optical flow approach, we perform two stream implementation where two encoders are used for each input type. The encoders share some of the final layers to reduce network size, and skip connections are passed from both streams.
Since our network does not have any fully connected layers or an extra region proposal network, the network has fewer number of parameters and can be trained end-to-end in a single stage. 


\paragraph{Optimization} We use Adam optimizer~\cite{kingma2014adam} with an initial learning rate of 1e-4 and finetune at a rate of 1e-5. For our joint training task, we can fit an effective batch size of 14 clips per iteration. The model is trained for 5 epochs with initial learning rate and fine-tuned for another 6 epochs.

\paragraph{Joint training} 
We train all three branches together with the loss weights assigned based on class distribution per task. For the A2D dataset, \textit{STU-Mask} detection is given the weight of 0.3, while both actor and action detection task is given weights of 1.3 (based on per class pixel distribution). 

\paragraph{STU-Mask} We input the \textit{STU-Mask} of size 4 x 56 x 56 for the action detection task, which helps to increase focus on the related pixels. For training, we use all actor regions from ground truth as the \textit{STU-Mask}. During inference, we use the \textit{STU-Mask} predicted by the \textit{STU-Mask} branch.

\begin{figure*}[t!]
\begin{center}
\includegraphics[width=1.\linewidth]{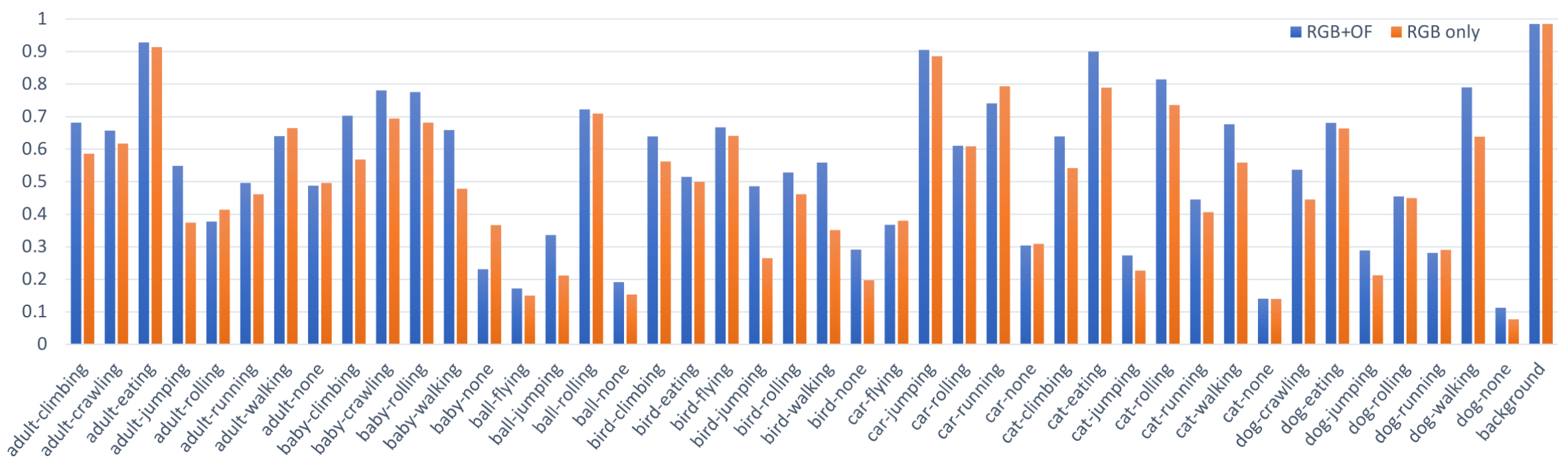}
\end{center}
\caption{Per actor-action category average accuracy score for A2D. The orange bars show results for RGB modality and blue bar for combined RGB and Optical Flow. We observe that on average, most of the classes benefit from having extra optical flow information.} 
\label{fig:a2d_avg_joint_acc}
\end{figure*}

\section{Experiments}
\subsection{Datasets}

\noindent
\textbf{A2D dataset}:
A2D \cite{A2DCVPR2015} is the first video dataset with multiple actor classes and action classes in the same clip along with semantic labels. It provides pixel-level semantic labels of 3-5 frames for each video and is the only joint actor-action segmentation benchmark reported in prior works~\cite{A2DCVPR2015, A2D2016GPM, A2D2017TSMT, A2D2018stanford}. The dataset consists of 3,782 YouTube videos, consisting of 7 actor classes performing one of the 9 action classes. A total of 43 actor-action pairs are valid and used for joint actor-action segmentation task. Both pixel level and bounding box annotation per actor-action pair are provided in this dataset.


\noindent
\textbf{VidOR dataset:}
We also evaluate our method on the VidOR dataset \cite{vidor2019annotating} which contains 10,000 videos with 80 object categories and 42 action predicates with bounding box annotations. Although it has more videos for training, the dataset is more challenging as it has a wide range of objects with a skewed distribution where ~92\% of objects are from only 30 categories. Each action is part of a triplet and consists of a subject and an object, with the subject performing the action. Thus, action detection using object and its surrounding context is more meaningful.

\subsection{Metric}
Following the evaluation protocols from~\cite{A2D2016GPM} and ~\cite{A2D2018stanford}, we measure average per-class accuracy $(ave)$, global pixel accuracy $(glo)$ and mean pixel Intersection-over-Union $(mIoU)$ as evaluation metrics. Accuracy is the percent of pixels with correct label prediction, where $(glo)$ is computed over all pixels and $(ave)$ is first computed per class and then averaged. Since background covers a large area and most models are biased towards background, mIoU is the most representative metric for correct pixel prediction over all classes~\cite{A2D2017TSMT}. We report results for actor, action and joint actor-action detection for $(ave)$, $(glo)$ and $(mIoU)$ metrics for a fair comparison with existing methods.

\begin{figure*}[t!]
\begin{center}
\includegraphics[width=1\linewidth]{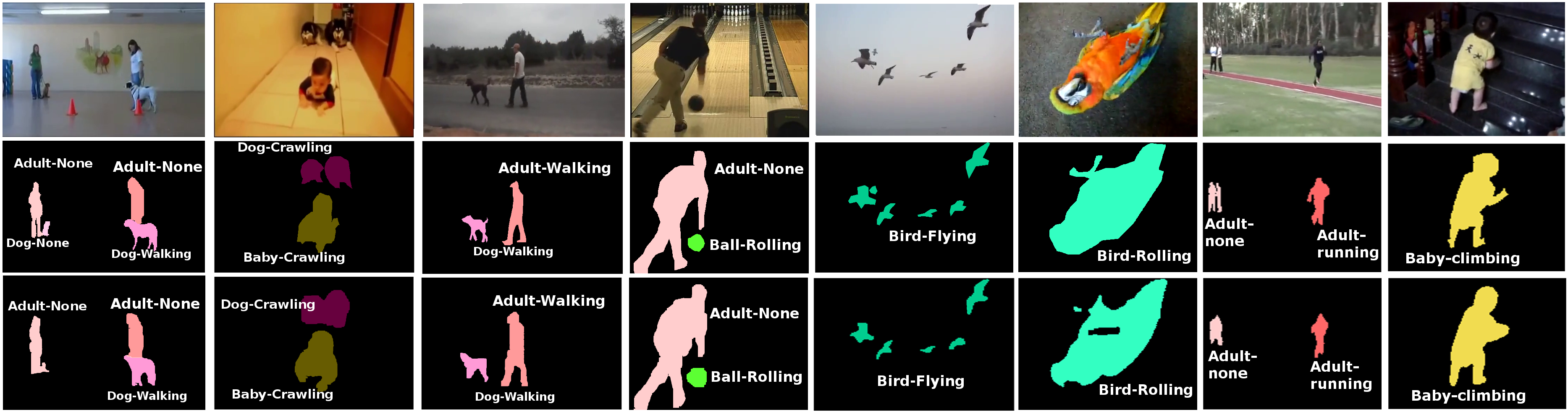}
\end{center}
\caption{Qualitative results of our method on A2D dataset. The first row shows the input key frame, second row shows the ground truth with annotation labels and third row shows our joint actor-action detection result with predicted labels.
}
\label{fig:a2d_eval}
\end{figure*}

\subsection{Results}
The performance of SSA2D on A2D is shown in Table~\ref{table_A2D_RGB}. Using only RGB stream, SSA2D gives improved joint actor-action mIoU with significant reduction in inference time \textit{($\sim$11x faster)}. This demonstrates that the network's joint training is able to learn action features based on actors while computing video level detection faster than previous methods. Moreover, using RGB+OF input we observe that the network gives improved mIoU scores on action and joint task as expected, demonstrating that our approach generalizes to different types on input modalities. We also analyze per class performance and the scores are shown for both of our RGB model and RGB+OF model in Figure~\ref{fig:a2d_avg_joint_acc}. We observe that the proposed method can detect most classes accurately in RGB model and the scores are further increased with additional flow information. 

We report the performance of proposed method on VidOR dataset in Table~\ref{table_A2D_ablation}. We evaluate the model on same evaluation metrics as used for the A2D dataset. Due to its long tail distribution, the dataset suffers from large data unbalance. As such, even when our network performs well on those classes, the average accuracy and mean IoU scores drop due to tail classes with fewer training samples.

\begin{figure}[t!]
\begin{center}
\includegraphics[width=0.45\textwidth]{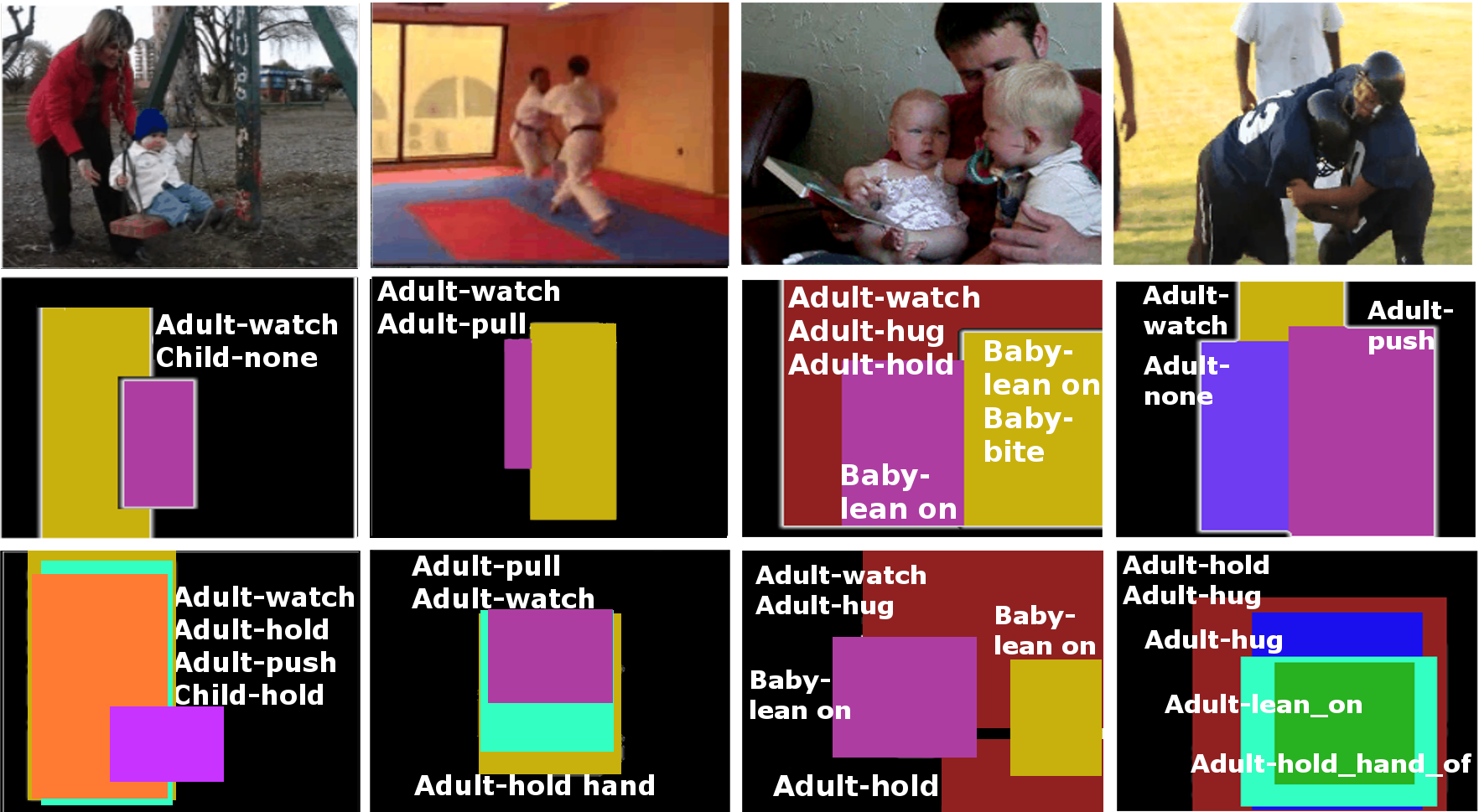}
\end{center}
\caption{Qualitative results of our method on ViDOR dataset. The top, middle and bottom row represents input key frame, ground truth and our joint actor-action predictions with label respectively.} 
\label{fig:vidor_samples}
\vspace{-1em}
\end{figure}

\begin{figure}[h]
\begin{center}
\includegraphics[width=0.45\textwidth, height=0.24\textwidth]{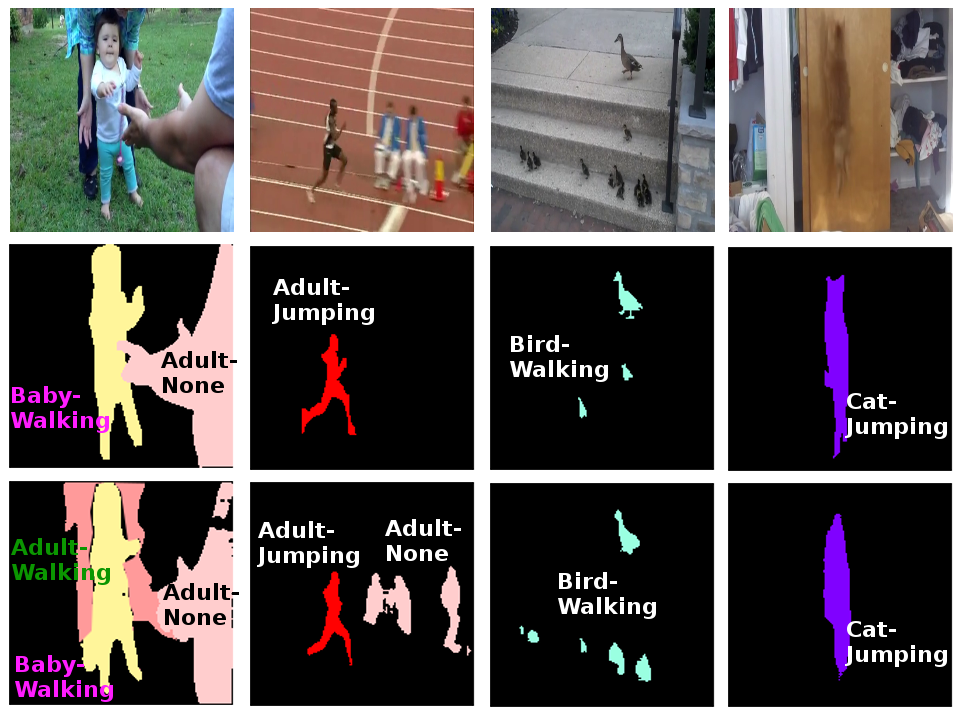}
\end{center}
\caption{Qualitative analysis of some success cases where the network predicts better than the ground truth. The top, middle and bottom row represents input key frame, ground truth and our joint actor-action detection predictions with label respectively. The network correctly predicts labels for actions which are not annotated in the ground truth but present in the clip. 
} 
\label{fig:success_cases}
\vspace{-1em}
\end{figure}

\noindent
\textbf{Qualitative evaluation}:
Figure \ref{fig:a2d_eval} and \ref{fig:vidor_samples} show qualitative results for actor-action detection. We observe that the proposed method can predict reasonable detections for most of the cases. Figure \ref{fig:success_cases} shows that the network predicts correctly even though ground truth annotation is missing labels. The network is able to generalize and learn effective actor-action features to predict the missing labels. The last column shows the network detecting a hard sample correctly. Even though the cat blends with the background, it is well segmented and detected as \textit{cat-jumping} class. Using 3D convolution on videos where the object is better visible in other frames, detection improves in such challenging frames as features is evaluated together for the entire video.




\subsection{Ablation Studies}
We further validate the importance of different components proposed in our model through ablation experiments. Since our contribution is agnostic to input type, we evaluate all variations against the full RGB only model in Table~\ref{table_A2D_ablation}. 


\textbf{Actor Prior Infusion (\textit{AP-Infusion})}:
One of the key components in improving action detection in our model is the use of actor prior for inferring activities. The \textit{A-prior} coming from actor detection branch provides contextual information regarding all actors around each pixel. It is reasonable to have an understanding of the actors involved in order to better judge the actions happening. While~\cite{gkioxari2017detecting} shows that using pair-wise actor features helps improve action classification in images, our \textit{AP-Infusion} approach uses all of the involved actor's features together because of the pixel-level detection. We train our model without using the \textit{AP-Infusion} to evaluate its effectiveness. As seen in \textit{Full} and \textit{w/o Actor-prior} models of Table~\ref{table_A2D_ablation}, \textit{A-prior} provides a significant gain in the action detection task ($\sim$ 6\% improvement in mIoU for A2D) and subsequently increases the scores for all other tasks. Since we perform a joint training, we also observe the decrease of scores for actor detection task when feedback from the \textit{AP-Infusion} block is not present.

\textbf{SSA-Masking}:
\textit{SSA-Masking} is used in action detection task to filter and enhance focus on action regions for pixel-wise detection. This reduces the surplus background noise and helps in a faster convergence. Our motivation to use the \textit{STU-Mask} is to provide emphasis on features related to actors while filtering out excess background data. In RPN based methods, ROI-Pooling play the role of feature filtering. However, pooling is performed for each proposal independently making it computationally expensive. We use a unified mask for all the actors in the scene for this filtering making SSA2D more efficient. SSA-Masking enables the network to focus more on the actor pixels while suppressing the background pixels, which leads to an improved network performance for action detection ($\sim$ 3\% increase in mIoU for A2D) and also provides a faster network convergence ($\sim$3x).

\subsection{Comparative analysis}
Figure~\ref{fig:intro} shows a comparative view of our method along with~\cite{A2D2018stanford, A2D2016GPM, dang2018actor} in terms of performance and speed. Compared to~\cite{dang2018actor}, our training \textbf{does not} use weights pre-trained on segmentation task and trains the decoders from scratch, while~\cite{dang2018actor} uses pre-trained weights on segmentation tasks.  We observe that our method performs significantly better compared to~\cite{A2D2018stanford, A2D2016GPM} in all evaluation metrics as seen in Table \ref{table_A2D_RGB}. We see that despite fast inference time for~\cite{A2D2018stanford}, it under-performs and has a larger model. Furthermore, our quantitative scores are similar or slightly better than previous state of the art method~\cite{dang2018actor} and has significantly better inference time($\sim$11x). This large gap in inference time makes our approach better suited for actor-action detection in videos as compared to all prior works. 

\paragraph{Network parameters:} Another key aspect of the proposed method is the smaller network size (35M params for RGB and 55M params for RGB+OF) compared to~\cite{dang2018actor, A2D2018stanford} (44M params for RGB and 88M params for RGB+OF). Compared to prior works, SSA2D has reduced network size which relates to the overall efficiency and performance speed. The memory-efficient reduced network also enables end-to-end training for all tasks simultaneously as compared to multi-stage training \cite{dang2018actor}, which is time consuming.

\paragraph{Running time:}
A crucial difference between SSA2D and prior works is that previous works rely on RPN as an auxiliary task during training to obtain actor regions for ROI pooling. Our method uses end-to-end pixel-wise detection and jointly trains actor-action tasks on pixel level while keeping the implementation efficient and effective. For a fair evaluation, we evaluate the time taken to perform the evaluations on a single core of an Intel Xeon 2.3GHz CPU using a single NVidia Tesla K80 GPU \cite{dang2018actor}. During inference, our system takes $180$ \textit{ms per frame} with the RGB + OF model, while it takes only $67$ \textit{ms per frame} for single stream RGB model. ~\cite{dang2018actor} report computational time of $1100$ \textit{ms per frame} for their full system with optical flow, with around $350$ \textit{ms} being used for optical flow estimation.


\section{Conclusions}

We propose SSA2D, a simple yet effective approach for single-shot actor-action detection in videos. We demonstrate that actor-action detection in videos can be performed without relying on region proposal network where thousand of proposals are required making it in-efficient for dense video scenes. We evaluate the proposed approach on A2D and VidOR datasets and achieve comparable (sometimes even better) performance when compared with prior works. The proposed model can be efficiently trained (2x faster) with a fast inference ($\sim$11x  faster for RGB and $\sim$6x faster for RGB+optical-flow) with fewer network parameters when compared with best performing prior works.


\paragraph{Acknowledgments}This research is based upon work supported by the Office of the Director of National Intelligence (ODNI), Intelligence Advanced Research Projects Activity (IARPA), via IARPA R\&D Contract No. D17PC00345. The views and conclusions contained herein are those of the authors and should not be interpreted as necessarily representing the official policies or endorsements, either expressed or implied, of the ODNI, IARPA, or the U.S. Government. The U.S. Government is authorized to reproduce and distribute reprints for Governmental purposes notwithstanding any copyright annotation thereon.

{\small
\bibliographystyle{ieee_fullname}
\bibliography{main}
}

\end{document}